\documentclass[10pt,twocolumn,letterpaper]{article}

\usepackage{iccv}              % To produce the CAMERA-READY version

\usepackage{amsmath}
\usepackage{mathrsfs}
\usepackage{setspace}
\usepackage{algorithm}
\usepackage{algorithmic}
\usepackage{graphicx}
\usepackage{array, multirow}        
\usepackage{graphicx}              
\usepackage{caption, subcaption}   
\usepackage[utf8]{inputenc}         
\usepackage[T1]{fontenc}            
\usepackage{url}                   
\usepackage{booktabs}
\usepackage{booktabs} % 用于美化表格线条
\usepackage{amsmath} % 用于数学符号
\usepackage{stfloats}
\definecolor{iccvblue}{rgb}{0.21,0.49,0.74}
\usepackage[pagebackref,breaklinks,colorlinks,allcolors=iccvblue]{hyperref}

\def\paperID{11158} % *** Enter the Paper ID here
\def\confName{ICCV}
\def\confYear{2025}

\title{SAM Encoder Breach by Adversarial Simplicial Complex Triggers Downstream Model Failures}

\author{Yi Qin$^{2,5*}$, Rui Wang$^{1,3,4,6}$\thanks{Equal contribution.}, Tao Huang$^{2,5}$, Tong Xiao$^{2,5}$, Liping Jing$^{1,2,5}$\thanks{Corresponding author.}\\
$^{1}$State Key Laboratory of Advanced Rail Autonomous Operation, Beijing, China;\\$^{2}$Beijing Key Lab of Traffic Data Mining and Embodied Intelligence;\\ $^{3}$Collaborative Innovation Center of Railway Traffic Safety, Beijing, China;\\$^{4}$National Engineering Research Center of Rail  Transportation  Operation and Control \\System, Beijing, China; $^{5}$School of Computer Science and Technology, Beijing Jiaotong University,\\Beijing, China; $^{6}$School of Automation and Intelligence, Beijing Jiaotong University, Beijing, China\\
{\tt\small {\{yeeqin,rui.wang,thuang,21271023,lpjing\}}@bjtu.edu.cn}
}
\begin{document}
\maketitle
\begin{abstract}

While the Segment Anything Model (SAM) transforms interactive segmentation with zero-shot abilities, its inherent vulnerabilities present a single-point risk, potentially leading to the failure of numerous downstream applications. Proactively evaluating these transferable vulnerabilities is thus imperative. Prior adversarial attacks on SAM often present limited transferability due to insufficient exploration of common weakness across domains. To address this, we propose Vertex-Refining Simplicial Complex Attack (VeSCA), a novel method that leverages only the encoder of SAM for generating transferable adversarial examples. Specifically, it achieves this by explicitly characterizing the shared vulnerable regions between SAM and downstream models through a parametric simplicial complex. Our goal is to identify such complexes within adversarially potent regions by iterative vertex-wise refinement. % and maximize its geometric coverage through iterative, vertex-wise refinement. %This complex, containing multiple adversarial simplices, is iteratively refined via a vertex-wise approach. %to maximize adversarial loss and its volume. To bridge domain divergence, we introduce a lightweight domain re-adaptation strategy that aligns target task data with SAM’s source domain using minimal reference data.  
A lightweight domain re-adaptation strategy is introduced to bridge domain divergence using minimal reference data during the initialization of simplicial complex. Ultimately, VeSCA generates consistently transferable adversarial examples through random simplicial complex sampling.
% Notably, VeSCA leverages only the encoder of SAM, which mitigates overfitting issue, and generates consistently transferable adversarial examples by random simplicial complex sampling. 
Extensive experiments demonstrate that VeSCA achieves  performance improved by 12.7\% compared to state-of-the-art methods across three downstream model categories across five domain-specific datasets. Our findings further highlight the downstream model risks posed by SAM’s vulnerabilities and emphasize the urgency of developing more robust foundation models. The implementation code is available at \href{https://github.com/Seveone/SAM_VeSCA.}{https://github.com/Seveone/SAM\_VeSCA}

\end{abstract} 

\label{sec:intro}
\begin{figure}[h]
    \centering

    \includegraphics[width=0.5\textwidth]{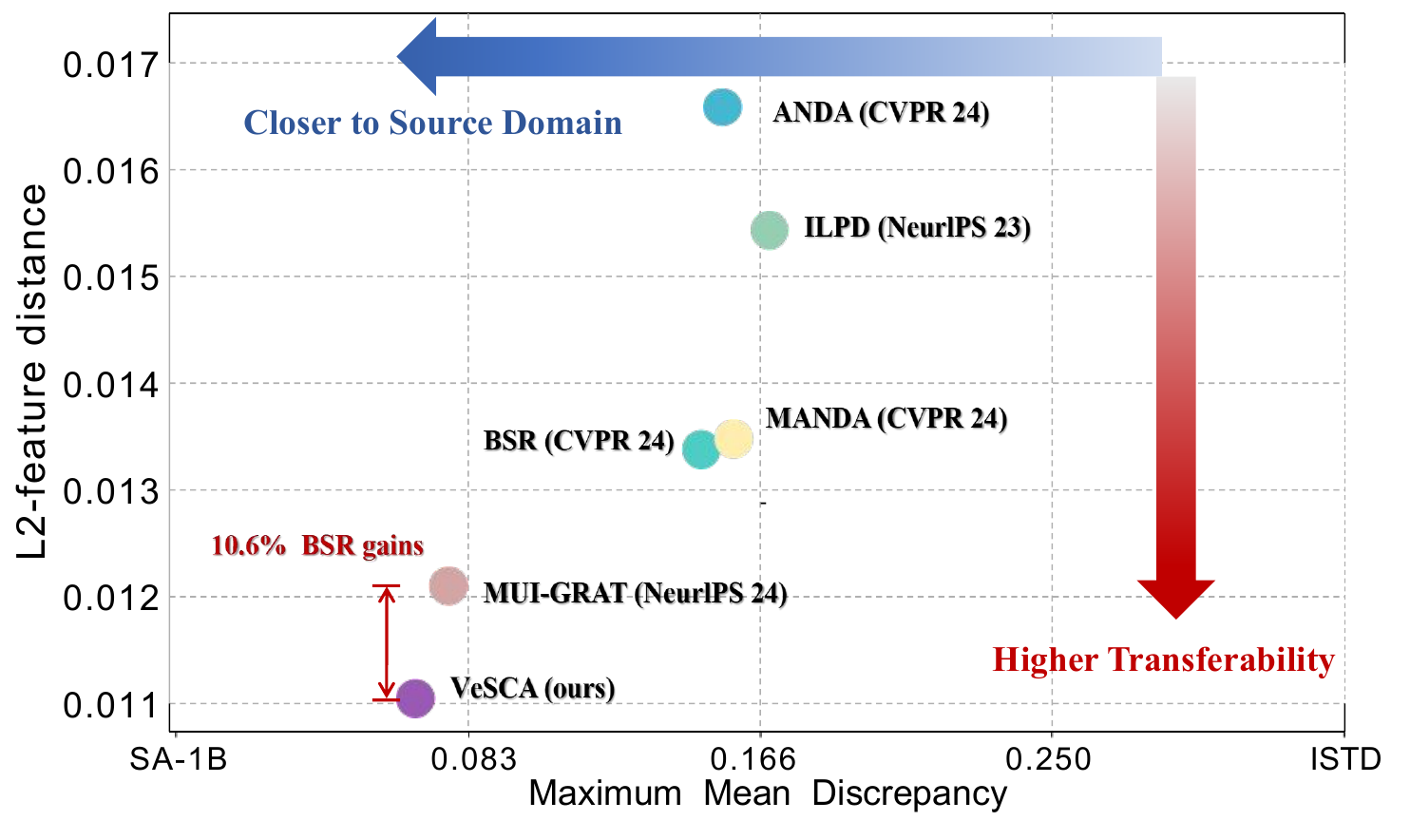} % 调整宽度以适应栏宽
    \caption{Performance comparison between VeSCA and baseline methods on the shadow segmentation task (ISTD ~\cite{wang2018stacked}). $y$-axis indicates the mean $\ell_2$ distance between features of adversarial examples extracted by SAM encoder and downstream model. $x$-axis shows Maximum Mean Discrepancy between adversarial examples and samples from SAM training data (SA-1B~\cite{kirillov2023segment}). The point at the lower-left corner indicate the ideal performance. }
    % \vspace{-0.5cm}
    \label{fig:teaser}
\end{figure}

%\vspace{-1.5cm}

\section{Introduction}
%Visual Foundation Models (VFMs) \cite{bommasani2021opportunities} have emerged in a variety of fields \cite{touvron2023llama,kirillov2023segment,chung2024scaling,brown2020language} which are pre-trained large models trained on large-scale datasets. Recently, Segment Anything Model (SAM)\cite{kirillov2023segment} which is interactive and capable of solving complex and comprehensive segmentation tasks.Leveraging its powerful zero-shot capability,has been quickly integrated into a wide range of downstream tasks\cite{ke2023segment,ren2024grounded,hetang2024segment,chen2023sam,mazurowski2023segment,zhang2023customized,wang2023samrs},such as road network graph prediction \cite{hetang2024segment},shadow segmentation \cite{chen2023sam},camouflaged object segmentation\cite{chen2023sam}, medical segmentation\cite{mazurowski2023segment,zhang2023customized}, and remote sensing image segmentation\cite{wang2023samrs}.

%These open-source pre-trained encoders or foundation models like SAM ,which possess strong feature extraction capabilities,are widely used in downstream tasks, also propagate their inherent security risks and vulnerabilities as they are increasingly deployed\cite{anderljung2023frontier}.Adversarial attack designed based on the instability of the algorithm \cite{dong2020adversarial}and the vulnerabilities of model structure can help to comprehensively explore the foundation model vulnerabilities and promote the development of new model defense designs.

The advent of foundation models has precipitated a paradigm shift in artificial intelligence, endowing machines with unprecedented generalization capabilities across diverse tasks. Models such as  CLIP~\cite{radfordClip2021}, DinoV2~\cite{oquab2023dinov2}, GPT4~\cite{achiam2023gpt}, DeepSeek-R1~\cite{guo2025deepseek}, Segment Anything Model (SAM)~\cite{kirillov2023segment} exhibit near-human proficiency in zero-shot perception and reasoning. Among these, SAM—a pioneering visual foundation model—has redefined interactive segmentation by dynamically generating masks in response to arbitrary prompts, such as points, boxes, or text. SAM has fostered applications in various domains, including shadow segmentation \cite{chen2023sam}, road network graph prediction \cite{hetang2024segment}, camouflaged object
segmentation \cite{chen2023sam},  medical segmentation \cite{mazurowski2023segment,zhang2023customized}, and remote sensing image segmentation\cite{wang2023samrs}. 

Yet, as an open-source foundation models, SAM embodies a double-edged sword~\cite{bommasani2021opportunities, anderljung2023frontier}: while its adaptability allows domain-specific tasks to achieve advanced visual comprehension even with limited data, the inherent vulnerabilities of SAM~\cite{xia2024transferable,zhou2025darksam} pose significant risks to downstream applications. A breach in SAM’s robustness could lead to systematic failures in adapted models fine-tuned for specialized tasks~\cite{xia2024transferable}. Thus, it is pivotal to explore these vulnerabilities proactively and alter the foundation model into a security choke point, as its robustness can also be inherited to the adapted tasks~\cite{Shafahi2020Adversarially}.

Early SAM attacks relied on white-box methods requiring precise prompt localization~\cite{huang2023robustness, zhang2023attack} or regional prompts~\cite{shenRegionlevelAttack2024}. Then, the research focus shifts to more practical threaten model and prompt-agnostic attacks targeting solely SAM image encoder~\cite{zheng2024PATA} is proposed, where adversarial examples maximize embedding divergence between clean images and their adversarial counterparts, but lack verified transferability to downstream models. Transferable attacks~\cite{madry2017towards,dong2018boosting,dong2019evading,wang2021enhancing,li2023improving,wang2024boosting,fang2024strong} exploit intrinsic vulnerabilities like non-robust features~\cite{tramer2017space}, architectural priors~\cite{dong2019evading,wu2020skip}, and algorithmic instability~\cite{zhangWhyDoesLittle2023,fang2024strong} to create cross-architecture perturbations. However, domain shifts with divergent data distributions weaken their efficacy. Recent UAP-based approaches~\cite{xia2024transferable,zhou2025darksam} mitigate this via source-domain perturbation learning. Nevertheless, these methods demand extensive data from source domain for perturbation learning and encounter limited downstream task transferability.

%To resolve these issues, in this paper, we propose Pointwise Augmentation Simplicial Complex Attacks(PASCA). By leveraging the inherent vulnerabilities of the open-source SAM visual encoder and the instability of training algorithms, we learn a adversarial simplicial complex to get infinite adversarial samples by multiple samplings from this adversarial subspace.Without accessing the downstream tasks and training datasets, they effectively misleading SAMs and various fine-tuned models, as shown in \Cref{fig:my_label}.
Existing research~\cite{tramer2017space,ma2018characterizing} reveals the existence of shared adversarial subspaces, which often locate on the higher-dimensional data manifolds. The transferability of adversarial examples originates from the overlap of high-dimensional adversarial subspaces~\cite{tramer2017space}. Inspired by this idea, we propose Vertex-Refining Simplicial Complex Attacks (VeSCA), a novel method that leverages a parametric subspace learning strategy to construct a adversarial subspace using only the encoder of the open-source SAM.
% VeSCA exploits the inherent vulnerabilities of the open-source SAM visual encoder for comprehensive robustness testing across various downstream models. 
Specifically, we explicitly model and learn an adversarial simplicial complex—referred to as the adversarial subspace~\cite{tramer2017space,maCharacterizingSubspaces2018}. Our approach guides the search by maximizing its volume within high-loss (adversarially potent) regions, allowing to comprehensively capture the vulnerability of the SAM visual encoder.
% Specifically, we identify non-robust, vulnerable regions—referred to as the adversarial subspace~\cite{tramer2017space,maCharacterizingSubspaces2018}—by explicitly modeling \textcolor{red}{and learning} a adversarial simplicial complex, and guiding the search by maximizing its volume within high-loss (adversarially potent) regions. 
%for enhancing the transferability of adversarial perturbations.
%This design choice is motivated by the fact that, 
Compared to other learnable parametric structures (e.g., Bézier curves~\cite{goodfellow2015harnessing},high-dimensional spheres~\cite{lubana2023mechanistic}), simplicial complexes~\cite{benton2021loss} provide more general and flexible geometric representations, making them particularly suitable for high-quality modeling of adversarial subspaces. On this basis, 
VeSCA leverages an ensemble of adversarial simplexes that captures different modes (local optima on loss surface) to form a adversarial simplicial complex. This approach effectively approximates Bayesian marginalization~\cite{Wilson2020} to guarantee performance gains.

To ensure potency over the entire subspace, the initial vertex for each simplex should be located at one mode. To achieve this, we introduce an efficient domain re-adaptation strategy that repositions downstream target data (e.g., ISTD~\cite{wang2018stacked}) back to the source domain knowledge of SAM (i.e., SA-1B~\cite{kirillov2023segment}) using only limited number of reference data points. Finally, robustness test examples are generated by randomly sampling from the adversarial subspace characterized by the simplicial complex. Our experiments, as shown in \Cref{fig:teaser}, demonstrate that VeSCA significantly improves attack performance and achieves a competitive domain re-adaption compared with the costly UAP-based state-of-the-art method. The main contributions of this paper are as follows:

\begin{itemize}
    \item We propose a novel approach to explicitly characterize the shared vulnerable subspace between SAM and its downstream task models using adversarial simplicial complexes, which ensures more stable transferability compared to single-point adversarial examples~\cite{xia2024transferable,wang2024boosting} and distribution-based adversarial subspace~\cite{fang2024strong}, thereby enhancing the attacking efficacy across different tasks.
    \item We introduce an efficient domain re-adaptation strategy by minimizing the embedding distance between source domain of SAM and target task domain, using only a few source-domain images (e.g, 40 data points in our experiment), avoiding the resource-intensive UAP process. %patch-level data augmentation technique is introduced for the ViT architecture to select reference points for 
    \item Extensive experiments on three categories of downstream models, covering the original, adapter-tuned, and full-parameter fine-tuned SAM encoder, demonstrate up to 12.7\% improvement in attack performance over SOTA methods across five domain specified datasets. Our results further validate the systemic risks posed by single-point failures in the open-source SAM model, highlighting the urgency for designing intrinsically robust deep learning models.
\end{itemize}

\section{Related Work}
\label{sec:formatting}
In this section, we present a brief review of previous research on transferable adversarial attack (\cref{sub2.2}), identified vulnerability of SAM (\cref{sub2.2}) and adversarial subspace (\cref{sub2.3}).
%\subsection{Adversarial attacks that explore vulnerabilities}
%In \cref{sub2.1}, we first give a summary of previous research related tocadversarial attack ,and \cref{sub2.2} introduces SAM and related works on its adversarial vulnerability. Finally \cref{sub2.3} introduces subspace learning and adversial subsapce.
%\subsection{Adversarial attacks that explore vulnerabilities}
\subsection{Transferable Adversarial Attacks}
\label{sub2.1}
%Adversarial attack designs have explored the vulnerabilities of model architectures and training algorithms. Formally, let $f$ be victim model and $\mathcal{L}$ be the loss function (e.g., the cross-entropy loss) that evaluates the quality of the model's prediction. Let $\mathcal{B}_{\epsilon}(x) = \left\{ x' : \left\| x' - x \right\|_{p} \leq \epsilon \right\}$ be an $\ell_{p}$-norm ball centered at the input $x$, where $\epsilon$ is perturbation bound. For each input $x$, the untargeted adversarial attacks aim to find an adversarial perturbation $\delta$ by solving:
Transferable adversarial attacks aim to generate perturbations, typically by crafting attacks using a surrogate model \( f \), which approximates the behavior of unknown victim models. In the context of untargeted adversarial attacks, formally, the goal is to find a perturbation \( \delta \) by solving:
\begin{equation}
\begin{aligned}
    \max_{x+\delta \in \mathcal{B}_{\epsilon}(x)} \mathcal{L}_{CE}(f(x + \delta), y),
\end{aligned}
\end{equation}
where \( \mathcal{L}_{CE}(f(x), y) \) is a loss function (e.g., cross-entropy), \( y \) is the true label of the input \( x \), and  \( \mathcal{B}_{\epsilon}(x) = \{ x' : \left\| x' - x \right\|_p \leq \epsilon \} \) represents an \( \ell_p \)-norm ball centered at \( x \) with perturbation bound \( \epsilon \).

%Transfer-based attacks exploit intrinsic vulnerabilities, including dependence on non-robust features, shared architectural priors, and algorithmic instability caused by suboptimal loss design or unstable optimization dynamics.
%~\cite{zhangWhyDoesLittle2023,fang2024strong}. Some works exploite the vulnerabilities of model architectures. Image scaling data augmentation attack(SIM) \cite{lin2019nesterov} was proposed due to the discovery that CNNs struggle to capture features at different scales.And TIM\cite{dong2019evading} was developed based on the translational invariance of CNNs.Moreover, Wang et al.\cite{wang2020unified}~proved that negative correlation between adversarial transferability and the interaction between adversarial perturbation patches. So data augmentation attacks such as blocks shuffle and rotation (BSR)\cite{wang2024boosting} were developed to disrupt the model's attention mechanism which capture relationships between image patches.Considering the skip connection structure in ResNet, SGM \cite{wu2020skip} was designed to use gradients from skip connections rather than residual blocks. The layered architecture of DNNs may also lead to security vulnerabilities in feature extraction within the feature latent space.The intermediate-layer attack methods like ILPD\cite{li2023improving}introduce a perturbation attenuation strategy in intermediate layers to disrupt feature space representations.
These attacks exploit intrinsic vulnerabilities in neural networks, including their dependence on non-robust features and shared architectural priors. For instance, image translation and scaling-driven attack methods~\cite{dong2019evading,lin2019nesterov} exploit the structural priors of CNN, i.e., translational invariance through gradient averaging over shifted inputs and scale sensitivity via multi-scale resizing, to enhance adversarial transferability. Moreover, Wang et al.~\cite{wang2024boosting} propose to disrupt attention mechanisms in vision transformer~\cite{dosovitskiyimage} (ViT)-based models by block shuffle and rotation, due to the negative correlation between adversarial transferability and interactions among perturbation patches~\cite{wang2020unified}. Architectural vulnerabilities also extend to particular structural design and intermediate layers. Skip Gradient Method (SGM)~\cite{wu2020skip} prioritizes gradients from the skip connections of residual modules, while Intermediate-Layer Perturbation Decay (ILPD)~\cite{li2023improving} attenuates perturbations in latent spaces to destabilize feature representations. 

Furthermore, researchers~\cite{dong2020adversarial,goodfellow2015harnessing} have shown that the steep loss surface near the input data, stemming from the interplay between suboptimal loss design and unstable optimization dynamics, makes models highly sensitive to small input changes. This insight  has inspired gradient optimization methods~\cite{dong2018boosting, wang2021enhancing, lin2019nesterov, fang2024strong}, which aim to identify better individual maxima of the loss function. Additionally, studies indicate that adversarial transferability is primarily driven by the overlapping adversarial subspaces among models with different architectures~\cite{maCharacterizingSubspaces2018, ilyasFeatures2019}. Therefore, Fang et al. propose modeling these subspaces using a Gaussian-based distribution, further enhancing the performance of transferable attacks.

%To further explore the steep regions of the loss function,   To address this limitations,
% However, their work was conducted under a query-based threat model, limiting its scalability in real-world applications.the stochastic adversarial attack MANDA\cite{fang2024strong} was developed, which samples the posterior distribution of perturbations along multiple trajectories to approximate a Bayesian posterior. In this paper, we focus on the vulnerabilities of these two aspects to design our optimization method.

%\subsection{Segment Anything Model(SAM) and Safety}
\subsection{Identified Vulnerability in SAM}
\label{sub2.2}
Open-sourced SAM performs label-free mask prediction tasks through diverse prompts and demonstrates zero-shot segmentation performance. It consists of three key components: a ViT image encoder, a prompt encoder, and a compact mask decoder. Inevitably, like other forms of neural networks, SAM also exhibits adversarial vulnerabilities~\cite{huang2023robustness, zhang2023attack}.
Early attempts to attack SAM have primarily focused on white-box approaches targeting the entire SAM’s architecture, which require precise knowledge of prompt locations~\cite{huang2023robustness, zhang2023attack} or regional prompt~\cite{shenRegionlevelAttack2024}, limiting their practical realization. This led researchers to shift towards prompt-agnostic attacks, which target the most inheritable component of SAM’s image encoder~\cite{zheng2024PATA}. This process can be formalized as
\begin{equation}
    \begin{aligned}
    \max_{x + \delta \in \mathcal{B}_{\epsilon}(x) } &\mathcal{L} \left( f_{\phi_{im}}(x), f_{\phi_{im}}(x + \delta) \right), 
%&\text{s.t. } \delta_s = \mathcal{AT}(f_{\phi_{im}}, x_\tau),  
    \label{3}
    \end{aligned}
\end{equation}
where $f_{\phi_{im}}$ is an image encoder, $\mathcal{L}$ represents similarity metric for image embeddings. These attacks generate adversarial examples by maximizing the embedding distance between legitimate inputs and their adversarial counterparts, but their transferability to downstream models can not be guaranteed and barely verified. 

Successful transferable attacks~\cite{madry2017towards,dong2018boosting,dong2019evading,wang2021enhancing,li2023improving,wang2024boosting,fang2024strong} exploit intrinsic model vulnerabilities, such as dependence on non-robust features (data-driven biases)~\cite{tramer2017space,moosaviUAP2017}, shared architectural priors (model design biases)~\cite{dong2019evading,wu2020skip}, and algorithmic instability~\cite{zhangWhyDoesLittle2023,fang2024strong}, to create perturbations that consistently transfer across architectures. Nevertheless, these attacks face challenges in transferring across domains with divergent data distributions, where the gap between source and target tasks greatly hinders the effectiveness of adversarial examples. Recent efforts~\cite{xia2024transferable,zhou2025darksam} have taken a step forward on this issue by leveraging universal adversarial perturbation (UAP) techniques. Nevertheless, these techniques face challenges such as multi-stage preprocessing procedure, heavy dependence on source-domain datasets for initial perturbation optimization, and limited  transferability to downstream models.

\subsection{Parametric Subspace Learning}\label{sub2.3}
Prior research on machine learning has investigated how the geometric properties of complex, high-dimensional loss landscapes impact model generalization~\cite{garipov2018loss,freeman2016topology,draxler2018essentially,hochreiter1994simplifying,fort2019deep}. %A widely accepted view is that local minima, also termed as modes, are interconnected via low-loss pathways, with abundant optimal solutions situated along these routes. These connecting paths rapidly evolve into subspaces with rich parameter structures such as mode connector\cite{garipov2018loss,lubana2023mechanistic} which is crucial for enabling fast ensemble learning to enhance the generalization.
%Based on this insight, Benton\cite{benton2021loss} demonstrated that there are low-loss multidimensional manifolds formed by modally connected simplicial complexes, connecting many independently trained models.This method significantly outperforms previous works, because their method is an approximation to a Bayesian model average. Notably, simplicial complex relies on point-to-point training and simplicial complex sampling for predictions, demanding a strong initial pre-trained model as the starting vertex.
The local minima (also termed modes) have been empirically shown to interconnect via low-loss pathways\cite{garipov2018loss}, with abundant optimal solutions situated along these routes. Studies expanded these paths into subspaces with diverse parametric structures, including Bézier
curves~\cite{garipov2018loss}, high-dimensional spheres~\cite{lubana2023mechanistic}, and simplices~\cite{wortsman2021learning}, forming rich types of mode connector. Effectively learning these subspaces of model parameters enables fast ensemble learning, there of enhancing the generalization.

Notably, Benton et al.~\cite {benton2021loss} employ a mixture of simplexes as a multimodal posterior. It is realized by ensembling  models that sampled from these simplexes to approximate the Bayesian model average, which further improves the prediction accuracy and calibration. The simplicial complex relies on a point-to-point refining process, requiring that the initial vertex of each simplex correspond to a solution within a single mode to ensure efficacy.

% As mentioned earlier, adversarial subspaces~\cite{szegedy2013intriguing,tramer2017space}, typically have higher dimensionality than the input data~\cite{maCharacterizingSubspaces2018}. This leads to significant overlap between models, enabling stronger adversarial transferability. In this paper, we adopt the simplicial complex to explicitly characterize the shared vulnerable region between SAM and its downstream task models, allowing comprehensive robustness testing of this foundation model. 

%Similar connected spaces known as adversarial subspaces also exist in the process of generating adversarial examples\cite{szegedy2013intriguing}. Additionally,~adversarial subspaces typically require higher-dimensional representations\cite{stephan2017stochastic} leading to significant overlap between different models, thereby enabling better adversarial transferability across models.  
%In this paper, we model the simplicial complex space of perturbations and maximize its volume in order to identify the adversarial subspace of the SAM visual encoder. During the process of vertex-wise solving for the maximal adversarial simplicial complex,~the quality of the fixed initial vertex affects the subsequent vertices of the simplicial complex~thereby affecting the entire subspace.

\section{Method}%Method
\label{sec:method}
% \subsection{Motivation}
\begin{figure*}[h]
\centering
\includegraphics[width=0.90\textwidth]{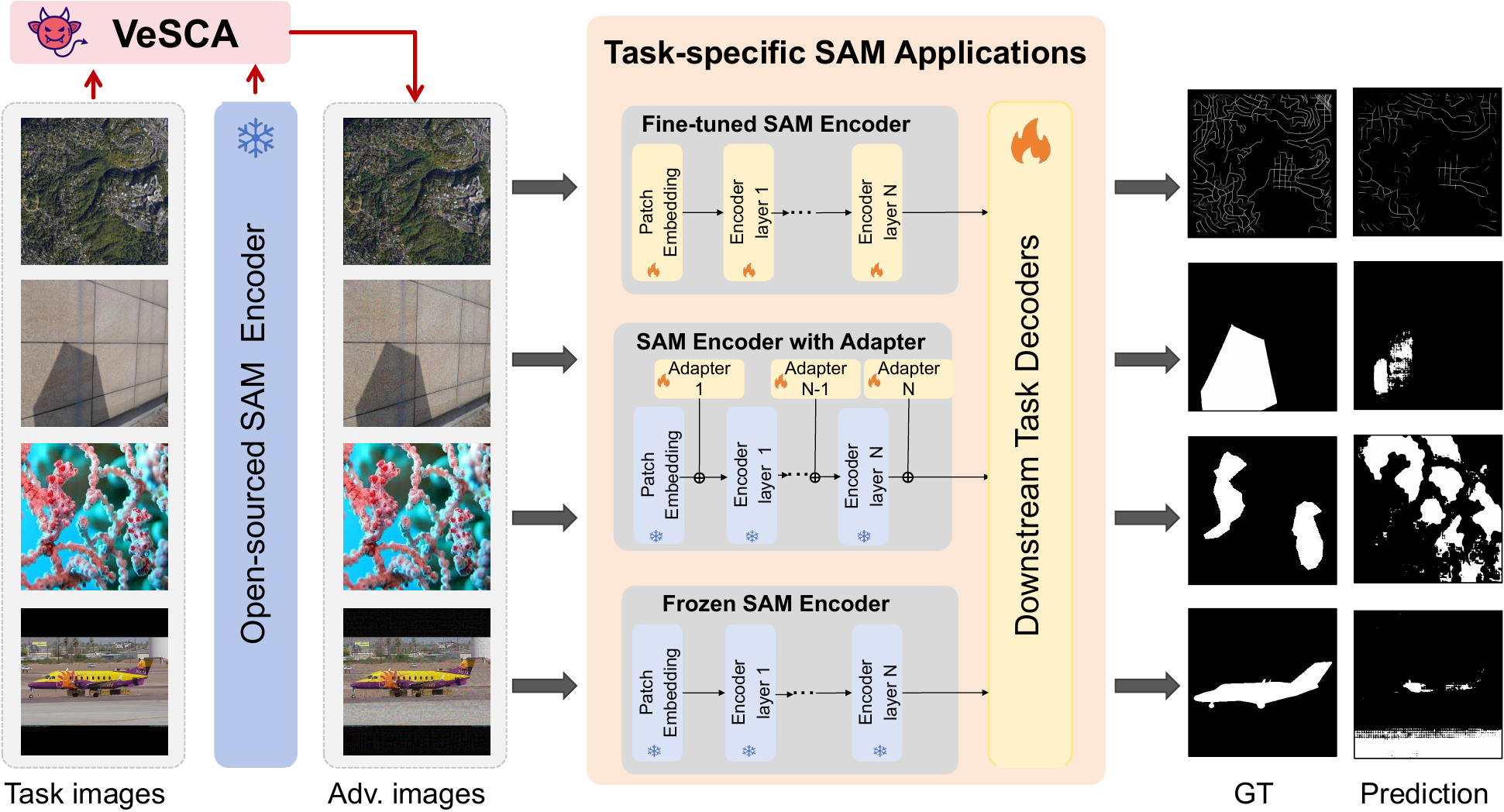}
\caption{Illustration of VeSCA framework for robustness testing of SAM downstream models. It achieves highly transferable attacks on various black-box tasks by exploring the vulnerability space of the open-source SAM encoder}
\label{fig:my_label}
\end{figure*}
In this section, we first reformulate the optimization objective in~\cref{3} to search for an adversarial simplicial complex (ASC) for each legitimate input  (\Cref{sec:pro}). Next, we propose a domain re-adaptation strategy to initialize the first vertex of each simplex, aligning target domain perturbations with SAM training data (\Cref{sec:att}). Finally, we introduce the VeSCA algorithm (\Cref{sec:algo}), which iteratively refines vertices of the ASC.

%\subsection{Adversarial Simplicial Complex Modeling}
%\subsection{Model Construction for ASC}
\subsection{Characterization for ASC}
\label{sec:pro}
% \vspace{-2mm}
As previously mentioned, SAM exhibits adversarial vulnerabilities, which can propagate to its downstream task models~\cite{xia2024transferable,zhou2025darksam}. Therefore, a highly transferable adversarial attack is essential for comprehensive robustness evaluation of downstream tasks. Existing studies on transferable adversarial subspaces emphasize that their high-dimensional nature increases the likelihood of intersection, leading to significant overlap across model-specific subspaces and thereby enhancing adversarial transferability.
Hence, to improve the transferability of adversarial examples generated from SAM to downstream models, we introduce a subspace learning strategy that efficiently explores adversarial subspaces by constructing and identifying a high-loss adversarial simplicial complex for each legitimate input, effectively capturing the vulnerabilities of the SAM model. 

In practice, the simplex offers several key advantages over other structures, such as Bézier curves and high-dimensional spheres. On the one hand, it provides superior representational capacity due to its flexibility.
On the other hand, sampling from the simplex is computationally efficient, as it can be generated using linear combinations of its vertices. These properties make it particularly suitable for high-quality modeling of adversarial subspaces.

Formally, we characterize the adversarial subspace as a simplicial complex with $N$ simplices is denoted as $\mathcal{K}(S_0, S_1, \dots,S_{N-1})$, where the $m$-th adversarial vertex in the $n$-th simplex $S_{n}$ is denoted as $x_{adv,m}^n$ (abbreviated as $x_{m}^n$). A simplex $S_{n}$ with $M$ vertices is defined as
\begin{equation}
S_{n} = \left\{ \sum_{m=0}^{M-1} \omega_m \cdot x_{m}^n : \omega_m \geq 0, \sum_{0}^{M-1} \omega_m = 1 \right\} \label{eq:10}
\end{equation}
% and $\mathcal{V}(S_{n})$ as the volume of the n-simplex.
%~Simplicial complexes with $N$ simplices are denoted $\mathcal{K}(S_0, S_1, \dots,S_{N-1}).~$

Thus, we reformulate the adversarial optimization objective~\cref{3} into 
\begin{equation}
    \begin{aligned}
    \max_{ \mathcal{K} }  \mathbb{E}_{x_{adv} \in \mathcal{K}}\mathcal{L} \left(f_{\phi_{im}}(x_\tau),~f_{\phi_{im}}( x_{adv})\right),  \label{eq:5}
    \end{aligned}
\end{equation}
where $x_\tau$ represents a clean image in the target dataset $\mathbb{D}_\tau$ for downstream task $\tau$. 
\subsection{Initialization for ASC} 
\label{sec:att}
As initial vertices critically influence the overall subspace quality in simplicial complex learning~\cite{benton2021loss}, they should precisely anchor simplices in regions of shared weakness. To this end, we adopt MIM\cite{dong2018boosting} to learn adversarial initial vertices, with a focus on the design of the optimization objective. A naïve design would maximize the distance between adversarial and original examples in the feature space to ensure the adversarial attack effectiveness of the initial vertices. However, misalignment between SAM encoder feature space and downstream models, due to divergent training data (SA-1B and task-specific datasets), causes inconsistent gradient directions during optimization~\cite{xia2024transferable}, severely limiting transferability.
 To bridge this gap, we propose a \textit{domain re-adaptation} approach (DRA) inspired by transfer learning~\cite{tzeng2014deep}, which aligns feature distributions of the source and target data-domains to better capture shared
model vulnerabilities. Specifically, we calculate the mean feature embedding across a random subset \(\mathcal{S}\) of SA-1B which is SAM's training data (with only 40 samples in our experiment) and minimize the distance between adversaries and the averaged feature.

The joint initialization objective is: 
%To align feature spaces and better capture the vulnerabilities of SAM, ~we introduce \textbf{domain re-adaptation} to reduce the distribution gap between downstream adversarial samples and the training data of SAM (SA-1B).~ Specifically, we used random subset $\mathcal{S} \subset \mathbb{D}_S$ of training dataset SA-1B to calculate the mean of their feature embeddings output by the surrogate model, and minimized the difference between the adversarial samples and this mean.~The optimization objective in the initial phase is as follows:
\begin{equation}
    \begin{aligned}
       & \max_{x_\tau + \delta  \in \mathcal{B}_{\epsilon}(x_\tau) }\mathcal{L}( f_{\phi_{im}}(x_\tau + \delta),f_{\phi_{im}}(x_\tau) )\\&-\mathcal{L} \left(  f_{\phi_{im}}(x_\tau + \delta),\overline{f_{\phi_{im}}(\mathcal{S})} \right), \label{eq:6}
    \end{aligned}
\end{equation}
where $\mathcal{L}$ is implemented with $\ell_1$-Loss, yielding the initial adversarial vertex $x_{0}^0$ as a reference for input $x_\tau$. The first component preserves attack potency through increasing the feature space distance between adversarial and clean samples, ensuring the perceptual relevance of perturbations. 
%The first component preserves attack potency through maximizing the $\ell_1$-distance between adversarial and clean samples in SAM's encoder feature space, ensuring the perceptual relevance of perturbations. 
Complementing this, the second term is the domain adaptation constraint, which reduces source-target domain discrepancies. %by minimizing $\ell_1$-distance between adversaries and the averaged feature of \(\mathcal{S}\). 
By jointly optimizing perturbations in vulnerable subspaces and integrating source domain characteristics, our method effectively enhances the cross-task transferability of ASC initial vertices. Refer to \cref{alg:searchmode} in Appendix for details. 
% Complementing this, the second term is a domain adaptation constraint inspired by transfer learning .It achieves this by aligning feature distributions of SAM's training data with those of downstream task features. By jointly optimizing perturbations in vulnerable subspaces and integrating source domain characteristics, our approach improves cross-task transferability. 

%\subsection{Adversarial simplicial complex algorithm}
\subsection{VeSCA Algorithm}
\label{sec:algo}
% \vspace{-1mm}
%\paragraph{ Optimization Objective and Volume Penalty }
%After obtaining an excellent initial simplex vertex, we construct the initial vertices of the entire simplicial complex through data augmentation. Considering the inherent vulnerabilities of SAM visual ViT-encoder architecture,we propose a patch-level data augmentation strategy to disrupt the self-attention mechanism of ViT, minimize the interaction between adversarial perturbation patches, and enhance the diversity of vertices.~Specifically, we define a segmentation scale $ns$, which indicates that the image is divided into $ns\times ns$ blocks. Each block is then rotated by a random angle $\theta$, reshaped to the original dimensions, and all blocks are rearranged to form an image with the same dimensions as the original.The entire i-th data augmentation process can be explicitly expressed as
After obtaining a ideal initial simplex vertex close to source domain, we construct the remaining vertices of the adversarial simplicial complex through vertex-wise refining approach.  Additionally, we introduce a patch-level augmentation strategy $\mathcal{PAR}$ targeting the inherent vulnerabilities of ViT-based encoder to construct $N$ adversarial simplices. This strategy disrupts the self-attention mechanism by suppressing spatial relation between adversarial patches. Specifically, we define a segmentation scale $ns$, which indicates that the image is divided into $ns\times ns$ patches. Each patch is then rotated by a random angle $\theta$, reshaped to the original dimensions, and all patches are rearranged to form a new image with the same original size. The entire $n$-th data augmentation process can be explicitly expressed as
\begin{equation}
     x^{n}_0 = \mathcal{PAR} (x^{0}_0,ns,\theta)~,~0 < n < N
\end{equation}
%Then we independently optimize each adversarial simplex to guarantee that the entire adversarial simplex is in a high-loss region (adversarial) and fully investigate the vulnerable boundaries. Specially,~first objective is to minimize the expected loss across the entire adversarial simplex, while the second is to maximize the volume of the adversarial simplex $\mathcal{V}(S_{n})$:

To investigate the vulnerable subspace within high-loss regions, we independently optimize each adversarial simplex to guarantee that the entire adversarial simplex is in a high-loss adversarial region and fully investigate the vulnerable boundaries (high volume). Specially, we design the adversarial simplex optimization objective to balance two criteria: (1) maximizing adversarial loss across the simplex and (2) expanding its geometric coverage via volume regularization $\mathcal{V}(S_{n})$. The joint objective is formulated as:
\begin{equation}
  \begin{aligned}
      \max_{S_n \subset\mathcal{B}_{\epsilon(x_{\tau})}}  \mathbb{E}_{x \in S_n}& \mathcal{L}(f_{\phi_{im}}(x), f_{\phi_{im}}(x_\tau) ) \\&+ \lambda\log(\mathcal{V}(S_{n}))
  \end{aligned}
  \label{eq:optsimplex}
\end{equation} 
% ensure that a random point  $x_{adv}$ (or its neighborhood) which sampled within adversarial simplex  $S_ i(x_0,x_1,...,x_{M-1})$  with M vertices is in a high-loss area and calculate the volume of the simplex as a second constraint term, ensuring that the resulting simplex has maximum volume:

%According to the definition of adversarial simplex in \cref{eq:10}, sampled $x_{adv}$ from $n$-th Simplex can be represented as a linear combination of endpoints in simplex and weight vector $\boldsymbol{\omega}$ follows a Dirichlet distribution \textbf{Dir}:
According to the definition of adversarial simplex in \cref{eq:10}, one adversarial samples $x_{adv}$ can be generated as a linear combination of vertices of $n$-th Simplex with weights $\boldsymbol{\omega}$ sampled from a Dirichlet distribution \textbf{Dir}:
\begin{equation}
    \begin{aligned}
            x_{adv} = \sum_{m=0}^{M-1} \omega_m \cdot x_{m}^n, ~~\boldsymbol{\omega} \sim \textbf{Dir}(\mathbf{1}) \label{eq8}
    \end{aligned}
\end{equation}
where $\mathbf{1}=(1,1, \dots,1)$ parametrizes the uniform  $\textbf{Dir}(\mathbf{1})$.
%We approximate the expectation termd in~\cref{eq:optsimplex}, we use \cref{eq8} to sample $H$ vertices from the known adversarial simplex for a Monte Carlo process to approximate the expectation, resulting in a new optimization objective:
We approximate the expectation termd in~\cref{eq:optsimplex} using $H$ Monte Carlo samples \(\{x_{\text{adv}_h}\}_{h=0}^{H-1}\) drawn from $\mathcal{V}(S_{n})$:
\begin{equation}
  \begin{aligned}
      \max_{S_n \subset \mathcal{B}_{\epsilon}(x_{\tau})}\frac{1}{H}&\sum_{h=0}^{H-1} \mathcal{L}(f_{\phi_{im}}(x_{adv_h}), f_{\phi_{im}}(x_\tau)) \\&+ \lambda\log(\mathcal{V}(S_{n})) ,  \label{eq9}
  \end{aligned}
\end{equation}
%The second constraint term is to maximize volume of simplex. As the order of the simplex increases, the volume of the simplex grows exponentially. Consequently, we define an adaptive regularization parameter $\lambda$ to ensure the stability of the training process. We use the Cayley-Menger \cite{sippl1986cayley} determinant to compute the volume of simplex. The C-M determinant is an efficient method for computation, especially when dealing with simplicial complexes with endpoints less than 5, where the computational complexity is relatively low. 
The simplex volume $\mathcal{V}(S_{n})$ in the second term of \cref{eq:optsimplex} grows exponentially with its dimensionality. To stabilize optimization process, we employ an adaptive $\lambda$ that scales inversely with $\mathcal{V}(S_{n})$. The volume is computed via the Cayley-Menger determinant~\cite{sippl1986cayley}.The C-M determinant is an efficient method for computation, especially when dealing with simplicial complexes with vertices less than 5, where the computational complexity is relatively low.  %, which provides efficient closed-form solutions for simplices with fewer than 5 endpoints.
% The volume Cayley-Menger \cite{sippl1986cayley} determinant is express as 
% \begin{equation}
%     \begin{aligned}
%         &CM(S_{i}) = \\& \left|
% \begin{array}{ccccc}
% 0 & d_{01}^2 & \cdots & d_{0(M-1)}^2& 1 \\
% d_{01}^2, & 0 & \cdots & d_{0(M-1)}^2& 1 \\
% \vdots & \vdots & \ddots & \vdots & \vdots \\
% d_{(M-1)0}^2&d_{(M-1)1}^2& \cdots & 0 & 1 \\
% 1 & 1 & 1 & 1 & 1 \\
% \end{array}  \label{de10}
% \right|.
%     \end{aligned}    
% \end{equation}
% where the shape of determinant is $(M+1)\times(M+1)$ and $d_{ij}$ represents the Euclidean distance between the $i$-th vertex and the $j$-th vertex.
Obtaining the volume from the CM determinant of simplex:
\begin{equation}
    \begin{aligned}
        \mathcal{V}(S_{n})^2 = \frac{(-1)^{M}}{((M-1)!)^2 2^{M-1}} CM(S_{n})  \label{v}
    \end{aligned}    
\end{equation}
Note that the volume calculation requires that the vertices of the simplex be no less than 3. When the vertices $M=3$, it degenerates into calculating the area of a triangle.
% \vspace{-1em}
% \paragraph{Vertex-Refining Attack Algorithm} 

To better solve the optimization problem formulated \cref{eq9}, we proposed the vertex-wise simplicial complex optimization strategy. Specifically, we get the initial vertex of adversarial simplex via \cref{alg:searchmode}. Then the algorithm fixes known points, initializes and optimizes a new point to added to each simplex. The optimization from the initial mode to the second vertex is special because it is impossible to calculate volume with two vertices. In this case, we only use the first term of \cref{eq9}. Starting from the second point, we incorporate the volume regularization term into the optimization. The initial position of the new point is set to be the centroid of the simplex, which is the average of the vertices of the simplex, thereby maintaining the stability of the simplex optimization. Each simplex is optimized to form the final simplicial complex $\mathcal{K}$
which has N simplexes, and each simplex is composed of M adversarial vertices. The final adversarial sample is sampled from $\mathcal{K}$. The detailed implementation is presented in \cref{alg:robust_attack} of Appendix.

\section{Experiment}
In this section, we validate VeSCA on four categories of downstream models and five datasets, compared with seven state-of-the-art baseline methods. 
%We also explore the impact of the domain re-adaptation strategy and PAR data augmentation.
\subsection{Experiment Setup}
The attack setting in our work involves perturbing only the SAM encoder, without access to the gradients of task-specific downstream models. Then the generated adversarial task images are fed into downstream models. The transferability of adversarial examples is evaluated by measuring performance changes in the downstream tasks. Our primary focus is to investigate how adversarial vulnerabilities in SAM affect various downstream task models. Since this setting is closely aligned with MUI-GRAT~\cite{xia2024transferable}, we follow its attack protocol and further extend it by introducing some downstream tasks. The detailed experimental setup is presented below.
\begin{table*}[htbp]

\footnotesize
\centering
% \caption{Comparison results of adversarial attacks on different models and the surrogate model is the open-sourced SAM. }
\label{excel1}
\begin{tabular}{@{}l|c|cc|cc|c|c@{}}
\toprule
\textbf{Model} & \textbf{Shadow-SAM} & \multicolumn{4}{c|}{\textbf{Camouflaged-SAM}} & \textbf{SAM-road} & \textbf{GroundedSAM} \\
\midrule
\textbf{Dataset}&\textbf{ISTD}&\multicolumn{2}{c|}{\textbf{CAMO}}&\multicolumn{2}{c|}{\textbf{CHAME}}&\textbf{CityScale}&\textbf{SeglnW}\\
\midrule
\textbf{Metric}&\textbf{BER (↑)}&\textbf{$S_\alpha$ (↓)}&\textbf{MAE (↑)}&\textbf{$S_\alpha$ (↓)}&\textbf{MAE (↑)}&\textbf{Rec~$\%$ (↓)}&\textbf{MAP~$\%$ (↓)}\\
\midrule
Clean\_Image & 1.011 & 0.847 & 0.07 & 0.896 & 0.033 & 67.913& 51.256 \\
\midrule
Attack-SAM & 1.433 & 0.658 & 0.102 & 0.615 & 0.117 & 47.147 & 46.391 \\
MIM & 3.008 & 0.372 & 0.249 & 0.342 & 0.223 & 34.146 & 42.106 \\
ILPD & 4.597 & 0.346 & \underline{0.312} & 0.338 & 0.295 & 30.129 & 40.554 \\
BSR & 4.987 & 0.42 & 0.187 & 0.363 & 0.194 & 28.620 & 36.135 \\
ANDA & 3.324 & 0.342 & 0.235 & 0.340 & 0.255 & 32.374 & 37.224 \\
MANDA& 4.257 & \underline{0.331} & 0.297 & 0.335 & 0.294 & 29.516 & \underline{32.263} \\
MUI-GRAT & \underline{11.960} & 0.335 & 0.307 & \underline{0.322} & \underline{0.326} &  \underline{22.964} & 32.839 \\
\midrule
VeSCA & \textbf{13.228} & \textbf{0.316} & \textbf{0.320} & \textbf{0.317} & \textbf{0.330} & \textbf{20.045} & \textbf{28.934} \\
  (\%)&10.60(↑) & 4.53(↓) & 2.56(↑) & 1.55(↓) &1.23(↑) &12.71(↓) & 10.32(↓) \\
\bottomrule
\end{tabular}
\caption{Comparison results of adversarial attacks on various downstream models. The surrogate model is the open-sourced SAM encoder. The first three rows represent the adopted downstream models, the datasets, and the evaluation metrics used to measure the attack performance on corresponding models and datasets.
%The first row represents the downstream models, the second row displays the datasets, and the third row presents the evaluation metrics used to measure the attack performance of the corresponding models on these datasets.
The last row presents the results and percentage improved in performance of VeSCA compared to the second-best results. The strongest attacks are highlighted in \textbf{bold}, and the data for the second-strongest attack is \underline{underlined}.}
\end{table*}
\paragraph{Downstream tasks}  Experiments on downstream models (corresponding dataset) cover shadow-SAM-adpter \cite{chen2023sam} (ISTD \cite{wang2018stacked}), camouflaged object segmentation SAM-adpter \cite{chen2023sam} (CHAMELEON \cite{skurowski2018animal}, CAMO \cite{le2019anabranch}), road network graph extraction SAM-Road \cite{hetang2024segment} (CityScale \cite{he2020sat2graph}), and zero-shot segmentation Grounded SAM \cite{ke2023segment} (SegInW \cite{zou2023generalized}). 
%评估指标
To evaluate the attack performance across different tasks, we use Balance Error Rate (BER) for shadow segmentation, Mean Absolute Error (MAE) and Structural Similarity ($\mathcal{S}_\alpha$) for camouflaged object segmentation, Recall (Rec) for road network graph extraction, and Mean Average Precision (MAP) for zero-shot segmentation. The introduction of these datasets are provided in Appendix
% \vspace{-1em}
\paragraph{Baseline methods} We conducted thorough comparisons with a careful selection of representative baseline methods. Due to the recent emergence of research on the adversarial transferability targeting SAM model, directly comparable work remains limited.
%Given the relative novelty of research on SAM’s transferability and robustness, directly comparable work is limited. 
Beyond the highly relevant downstream transfer attack MUI-GRAT~\cite{xia2024transferable},we included Attack-SAM~\cite{chen2023sam} to specifically assess white-box SAM attacks. Furthermore,we incorporated well-established transferability attack methods primarily developed for visual classifiers. These include the classic first-order gradient-based attack MIM \cite{dong2018boosting}, input augmentation based attack BSR \cite{wang2024boosting}, intermediate-level feature based attack ILPD~\cite{li2023improving} to compare intermediate-layer feature attacks with our encoder’s feature-space distance optimization, SOTA adversarial subspace based attack ANDA and its advanced version MANDA~\cite{fang2024strong}.
\paragraph{Implementation details} We set the all baseline with the $ l_{\infty} $ bound $\epsilon= 10$, in our work attack update for each vertex in complex iterations $T=10$, uses only number of simplexes $N=4$, number of vertices $M=4$, and regularization parameter $\lambda^*=0.1$. For our Simplex Initialization, we set the number of mode's optimization iteration $t=10$, number of segmentation scale $ns=64$.  The hyperparameters listed above represent the current best settings. A more detailed analysis of parameter experiments is provided in the Section 4 of Appendix.

\subsection{Attack Performance} 
Main results, shown in \cref{excel1}, demonstrate that our proposed method achieves SOTA performance across all tasks, generating adversarial examples with stronger transferability and consistently posing significant adversarial threats to various downstream models.
Notably, VeSCA excels particularly on datasets with significant distribution differences: ISTD (shadow segmentation), CityScale (high-altitude satellite image), and SegInW (zero-shot segmentation). In these datasets, VeSCA's performance improves by more than 10\% compared to the second-best methods, with the most notable gain observed on the road network graph extraction model SAM-Road from the CityScale dataset, where it achieves a 12.71\% improvement. 

In the camouflaged object segmentation task, which includes the CAMO and CHAMELEON datasets, all encoder-based attack methods (except for Attack-SAM, a white-box attack against the full SAM model) demonstrated good transferability, with relatively small performance differences. This is likely due to the camouflaged objects in these datasets share a similar distribution with the SA-1B dataset used to train SAM. Nevertheless, VeSCA still achieves the best performance. The white-box attack method Attack-SAM underperforms other baselines, indicating that adversarial examples generated using the entire SAM model tend to overfit and thus exhibit weaker transferability to downstream tasks.

Moreover, we further investigated adversarial performance on the zero-shot task model Grounded-SAM using SegInW dataset. Even when the attacked images were unseen by the surrogate model, our method still achieved the best performance (28.934\%). This indicates that as long as SAM is used as part of the surrogate model within an AI system, our method can successfully launch effective attacks using only the downstream task dataset, without requiring any access to the downstream task model itself.
% In camouflaged object segmentation, apart from Attack-SAM, all encoder-based attack methods demonstrated strong transferability; however, VeSCA still achieved the best performance. The white-box attack Attack-SAM underperforms compared to other baselines, indicating overfitting and poor transferability of adversarial samples generated using the entire SAM. We also explored the attack performance on the zero-shot task model Grounded-SAM. Our method achieved SOTA performance (28.934\%), even when generating adversarial examples for images that the surrogate model had never seen before. This demonstrates that our approach, by leveraging only the SAM surrogate model, can effectively attacking without requiring downstream task information. 

% \vspace{-1em}
\subsection{Ablation Study}
% In this section, we analyze the impact on our method for initializing the simplicial complex on the ISTD dataset of Shadow-SAM, as shown in \cref{table2}.
In this section, we conduct ablation experiments on the shadow segmentation task with ISTD. 
%we conduct ablation experiments on the shadow detection model Shadow-SAM using the ISTD dataset. 
The study focuses on three aspects: (1) the impact of sample size in the domain re-adaptation strategy on the performance of VeSCA, (2) the contribution of each component within VeSCA, and (3) the effectiveness of adversarial subspaces modeling method with SCs, especially compared with the ones characterized using multivariate Gaussian distributions~\cite{fang2024strong}.
%a comparison of overall performance against MANDA, which models the adversarial subspace using multivariate Gaussian distributions.

\begin{figure}[htbp]
    \centering
    \includegraphics[width=0.95\columnwidth]{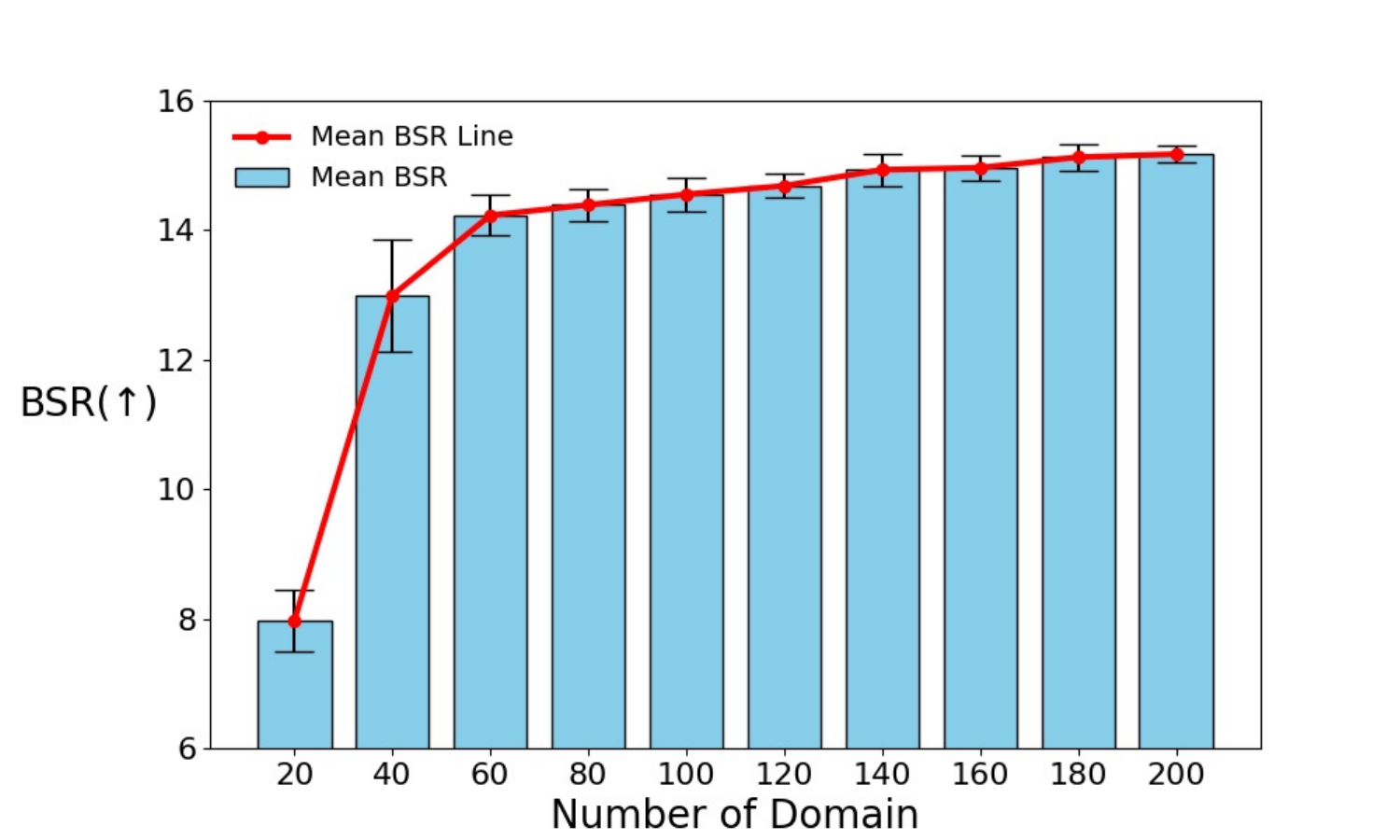}
    \caption{VeSCA performance with varied source-domain sample sizes during domain re-adaptation initialization on shadow segmentation task}
    \label{bsr_num}
\end{figure}

We first analyze the impact on the number of data points from SA-1B used in the initialization process for domain re-adaptation. We randomly sample between 20 and 200 images from the SA-1B dataset to construct source-domain subsets. For each subset size, the experiment is repeated five times, as shown in~\Cref{bsr_num}. The results indicate that the transferability of adversarial examples to downstream tasks stabilizes when more than 40 randomly selected source-domain images are used, and the performance surpasses that of the baseline methods.
% \begin{table}[htbp]
% \small
% \centering
% %\caption{\footnotesize Comparison on the ISTD under different domain sizes}
% % \caption{\footnotesize BSR change under different domain sizes}
% % \footnotesize  % 比\footnotesize更小的字号
% \begin{tabular}{|c|c||c|c|}
% \hline
% Number & BER (↑) & Number & BER (↑) \\
% \hline
% 20  & 7.97 $\pm$ 0.47  & 120 & 14.69 $\pm$ 0.18 \\
% 40  & 12.99 $\pm$ 0.86 & 140 & 14.93 $\pm$ 0.24 \\
% 60  & 14.23 $\pm$ 0.32 & 160 & 14.96 $\pm$ 0.19 \\
% 80  & 14.39 $\pm$ 0.24 & 180 & 15.13 $\pm$ 0.21 \\
% 100 & 14.55 $\pm$ 0.25 & 200 & 15.17 $\pm$ 0.13 \\
% \hline
% \end{tabular}
% \caption{VeSCA performance with different source-domain subset sizes during domain re-adaptation initialization on shadow segmentation task}
% \label{bsr_num}
% \end{table}

Next, we analyze the contribution of each component within VeSCA, as shown in \cref{abl}. Without domain re-adaptation (DRA), our data augmentation method PAR during the initialization phase, outperforms the current advanced BSR method and the traditional DTSI. Incorporating DRA as a constraint during the initialization phase significantly enhances performance. With the same BSR data augmentation strategy, adding the DRA constraint increases the BSE metric from 6.012 to 11.517, nearly doubling.
Additionally, we compared our method with MUI-GRAT \cite{xia2024transferable} during the initialization phase, replacing the adversarial optimization process with their well-trained universal perturbations (MUI). The experimental result shows that, under the same attack settings, adversarial examples generated from data distributions closer to SAM’s training domain exhibit stronger transferability to downstream tasks. This indicates that, without domain adaptation, there is no guarantee that downstream data with significant distributional differences from SA-1B would lead to correct optimization directions on SAM's encoder. Moreover, the results show that even with the  basic MIM, adding domain adaptation outperforms MUI, achieving higher performance (11.517). 
\begin{table}[htbp]
% \caption{BER($\uparrow$) on shadow segmentation task with adversarial examples sampled in Adversial Simplicial Complex (ASC) initialized by various data augmentations, domain re-adaption or MUI.}
%On shadow segmentation task, the BER($\uparrow$) of adversarial examples simpled within Adversial Simplicial Complex(ASC) initialized by different data augmentations,~doman re-adaption or MUI.
\small
\centering
\begin{tabular}{@{}c|c|c@{}}
\toprule
Type & Attack&BER (↑) \\
\hline
\multirow{3}{*}{Augmentation}
&ASC+DTSI&5.231\\
&ASC+BSR&6.012\\
&ASC+PAR(ours)&6.258\\
\hline
\multirow{2}{*}
% $\mathcal{L}_{\text{Domain}}$
{Initialization}&ASC+MUI&10.763\\
&ASC+DRA(ours)&11.517\\
\hline
Both&ASC+DRA+PAR&\textbf{13.228}\\
\bottomrule
\end{tabular}
\caption{BER($\uparrow$) on shadow segmentation task with adversarial examples sampled in Adversial Simplicial Complex (ASC) initialized by various data augmentations, domain re-adaption or MUI.}
\label{abl}
\end{table}
\begin{figure}[htbp]  % 使用浮动环境 figure
    \centering  % 图片居中
    \includegraphics[width=0.46\textwidth]{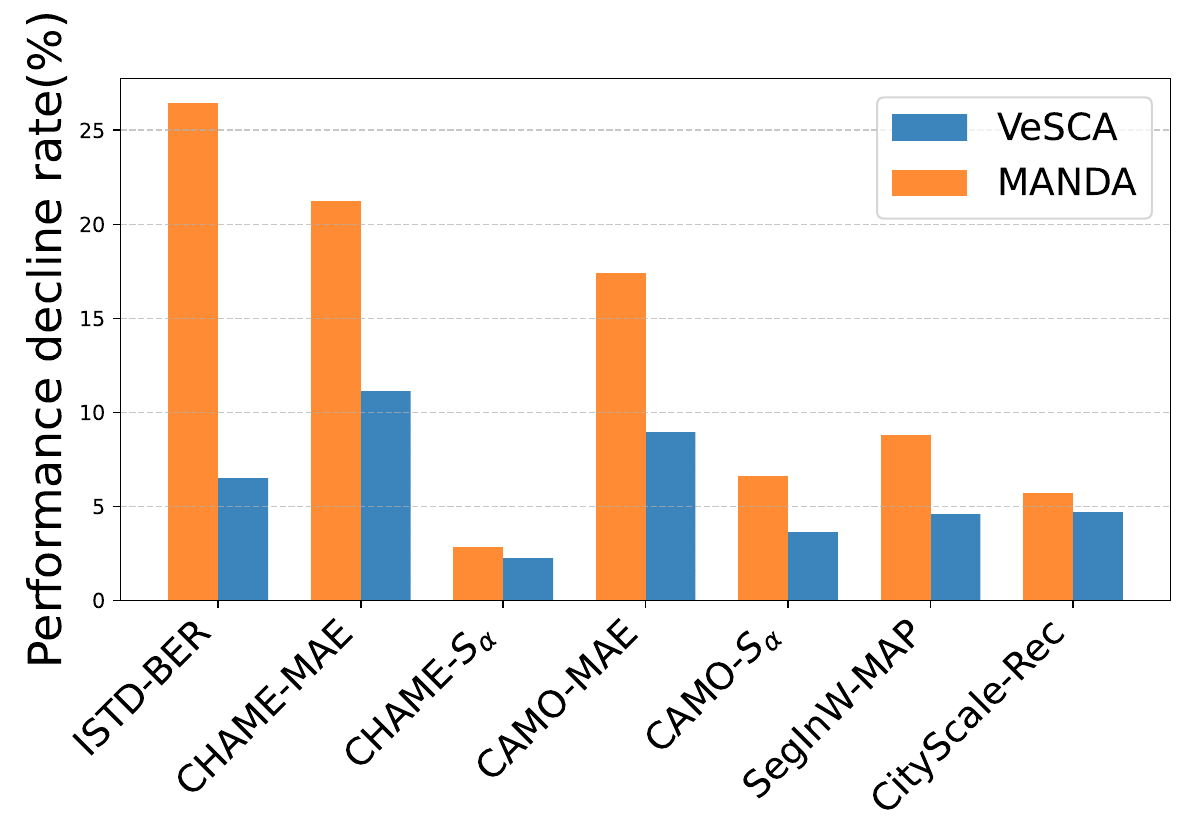}  % 插入图片，指定宽度为文本宽度的50%
    \caption{Performance Decline Rate (\%) illustrates the gap of
BER between adversarial subspace's center point and random points.~Adversarial subspace generated by VeSCA and MANDA with SAM-encoder surrogate model on the 5 datasets. }  % 图片标题
    \label{fig:sample}  % 图片标签，用于引用
\end{figure}

Additionally, to validate the effectiveness of different adversarial subspace modeling methods, we compared VeSCA with the distribution-based adversarial subspace method MANDA across all downstream tasks and various evaluation metrics. We measured the average performance gap between adversarial samples generated from the geometric center of the subspace and those from five random spatial sampling points. As shown in \cref{fig:sample}, on the CHAME dataset, the maximum performance gap of our method is about 11\%, while most other metrics show gaps below 10\%. Overall, the performance gaps between random samples and the geometric center in all tasks are smaller for VeSCA compared to MANDA. This indicates that VeSCA learns a flatter adversarial subspace than MANDA, resulting in more stable transferability of adversarial samples from random sampling. The experiments also confirm that VeSCA’s adversarial simplicial complex better characterizes the adversarial subspace than MANDA’s distribution-based approach, leading to stronger overall attack transferability—not just good performance at the center point. This clearly reveals the security risk inheritance caused by the shared vulnerable space between the SAM foundation model and its downstream task visual encoders. 
These findings underscore the critical need to strengthen the adversarial robustness for foundation models like SAM , thereby preventing a single foundation model from threatening its entire downstream ecosystem.
%This call for strengthened adversarial robustness for foundation models like SAM to prevent a single foundation model from threatening its entire downstream ecosystem.
\subsection{Time efficiency of computation}
\begin{table}[ht]
\centering
\label{tim}
\footnotesize
\vspace{-2mm}
% \scriptsize
% \setlength{\tabcolsep}{3.5pt} % 列间距压缩至3.5pt
\resizebox{\columnwidth}{!}{  % 强制表格宽度限制在半栏
\begin{tabular}{@{}l*{7}{c}@{}}
\toprule
\textbf{Method}       & MIM & ANDA & ILPD & BSR & Atk-SAM & M-GRAT & Ours \\
\midrule
\textbf{Time (s)} &
$\begin{array}{@{}c@{}} 0.37 \\ \pm 0.14 \end{array}$ &
$\begin{array}{@{}c@{}} 1.36 \\ \pm 0.11 \end{array}$ &
$\begin{array}{@{}c@{}} 0.81 \\ \pm 0.05 \end{array}$ &
$\begin{array}{@{}c@{}} 3.12 \\ \pm 1.05 \end{array}$ &
$\begin{array}{@{}c@{}} 0.65 \\ \pm 0.08 \end{array}$ &
$\begin{array}{@{}c@{}} 0.48^*\\ \pm 0.12 \end{array}$ &
$\begin{array}{@{}c@{}} 1.07 \\ \pm 0.13 \end{array}$ \\
\bottomrule
\end{tabular}}
\\
\scriptsize * Additional required UAP learning time is not included. %* represents additional UAP learning time required
\vspace{-1em}
\caption{Average time assumption per image-iteration}
\end{table}
The computational cost of VeSCA depends mostly on optimizing SCs and calculating their volume. However, configuring N=4 simplices with M=4 vertices yields optimal performance (see Fig. 2 in Appendix).
Although the overall time consumption is higher than other baelines, each of the $N$ adversial simplices is optimized independently. This allow us to use N GPUs for distributed computation. As a result, the optimization overhead is significantly reduced. The per-GPU experimental time is comparable to other methods in \cref{tim} (same experimental settings as \cite{xia2024transferable}). Furthermore, the offline robustness testing is less sensitive to time cost as stronger AEs enabling a more thorough vulnerability exploration.
% As shown in the \Cref{table2}, we finds that the impact of $\mathcal{L}_{domain} $ is huge, without using this constraint, the performance drops by nearly 53.7\%. We speculate that this is because when samples approach the training domain of the surrogate model, their optimization gradient can better approximate the gradient direction within the target model.In addition, during the initial simplex optimization process, the patch-level data augmentation module not only enhances the diversity of endpoints but also effectively targets the SAM encoder, outperforming better than SOTA data augmentation BSR.

\section{Conclusion}

% This work addresses the adversarial vulnerabilities of the Segment Anything Model (SAM) and its downstream applications by proposing Vertex-Refining Simplicial Complex Attack (VeSCA).
This work provides an adversarial robustness testing method regarding to the Segment Anything Model (SAM) and its downstream applications by proposing Vertex-Refining Simplicial Complex Attack (VeSCA).
VeSCA identifies shared adversarial subspaces between SAM and downstream models through parametric simplicial complexes, combining iterative vertex refinement and volume maximization to holistically encapsulate high-loss regions. A lightweight domain re-adaptation strategy aligns adversarial samples with training data distribution of SAM using minimal reference data, bypassing costly universal adversarial perturbation methods. Extensive experiments across diverse downstream tasks demonstrate the superior transferability of the proposed method, outperforming state-of-the-art attacks by up to 12.71\% in performance increasing rate. The geometric stability of the generated adversarial simplicial complexes underscores their fidelity to SAM’s latent space, highlighting systemic risks in open-source foundation models. This work urges prioritizing robustness in foundational AI systems while offering a practical framework for adversarial testing.

% In this paper, we introduce a novel black-box transfer attack, the Vertex-refining Simplicial Complex Attack(VeSCA), aimed at identifying highly transferable adversarial simplicial complex (ASC).
% Within the simplicial complex initialization strategy, VeSCA uses domain re-adaptation to closely align adversarial samples with the training data distribution, enhancing their cross-domain attack capabilities. It also leverages patch-level data augmentation to pinpoint an initial simplicial complex with substantial attack potential. The optimization approach incorporates vertex-wise refinement and volume maximization regularization, significantly improving the transferability of adversarial attacks.
% Additionally, we conducted experimental studies on four downstream task models across a total of five datasets, covering various downstream model architectures and tasks. We compared VeSCA with state-of-the-art methods. The results demonstrate that the ASC identified by VeSCA indeed closely resembles the training data distribution. The overall quality of the ASC is stable, with minimal performance difference between random sampling and the geometric center point, showing superior transferable attack capabilities over baseline methods.
\section*{Acknowledgements}

This work was supported by the National Key Research and Development Program of China under Grant 2024YFE0202900, the National Natural Science Foundation of China under Grants 62436001, 62406021 and 62176020, the State Key Laboratory of Rail Traffic Control and Safety, Beijing Jiaotong University (Contract No. RCS2023K006), the Joint Foundation of Ministry of Education for Innovation team under Grant 8091B042235, and the Fundamental Research Funds for the Central Universities under Grant 2025JBMC040 and 2019JBZ110, and the Fundamental Research Funds for the Central Universities (Science and technology leading talent team project) under Grant 2025JBXT006.

%Rui Wang reports financial support was provided by National Natural Science Foundation of China under Grant No.62406021. Liping Jing reports financial support was provided by National Key Research and Development Program under Grant No. 2024YFE0202900; National Natural Science Foundation of China under Grant Nos. 62436001, 62176020; the Joint Foundation of Ministry of Education for Innovation team under Grant No. 8091B042235; and the State Key Laboratory of Rail Traffic Control and Safety (Contract No. RCS2023K006), Beijing Jiaotong University.
% \input{ICCV2025-ESPRO/sec/X_suppl}
% \input{ICCV2025-ESPRO/rebuttal}
% \clearpage
{
    \small
    \bibliographystyle{ieeenat_fullname}
    \bibliography{main}
}
\end{document}